\documentclass{article}

\usepackage[english]{babel}

\usepackage[letterpaper,top=2cm,bottom=2cm,left=3cm,right=3cm,marginparwidth=1.75cm]{geometry}

\usepackage{amsmath}
\usepackage{graphicx}
\usepackage[dvipsnames]{xcolor}

\usepackage[colorlinks=true, allcolors=blue]{hyperref}
\newif\ifshowcomments
 \showcommentstrue 

\ifshowcomments

\newcommand{\michael}[1]{{\color{Blue}\sf{[Michael: #1]}}}
\newcommand{\michaeladdressed}[1]{{\color{cyan}\sf{[Michael: #1]}}}
\newcommand{\yaoqing}[1]{\textcolor{BurntOrange}{\sf[Yaoqing:\ #1]}}
\newcommand{\talita}[1]{\textcolor{green}{\sf[Talita:\ #1]}}
\newcommand{\gunther}[1]{\textcolor{OrangeRed}{\sf[Gunther:\ #1]}}

\newcommand{\guntheraddressed}[1]{\textcolor{blue}{\sf[Gunther:\ #1]}}
\newcommand{\tiankai}[1]{\textcolor{ForestGreen}{\sf[Tiankai:\ #1]}}
\newcommand{\jiaqing}[1]{\textcolor{orange}{\sf[Jiaqing:\ #1]}}
\newcommand{\caleb}[1]{\textcolor{teal}{\sf[Caleb:\ #1]}}

\newcommand{\geshi}[1]{\textcolor{violet}{\sf[GeShi:\ #1]}}
\newcommand{\ross}[1]{\textcolor{VioletRed}{\sf[Ross:\ #1]}}

\else

\newcommand {\michael}[1]{}
\newcommand{\michaeladdressed}[1]{}
\newcommand{\yaoqing}[1]{}
\newcommand{\talita}[1]{}
\newcommand{\gunther}[1]{}

\newcommand{\guntheraddressed}[1]{}
\newcommand{\tiankai}[1]{}
\newcommand{\jiaqing}[1]{}
\newcommand{\caleb}[1]{}

\newcommand{\geshi}[1]{}
\newcommand{\ross}[1]{}

\fi

\newcommand\SYSNAME{LossLens}

\title{\SYSNAME: Diagnostics for Machine Learning \\ through Loss Landscape Visual Analytics~\footnote{This work has been accepted by IEEE Computer Graphics \& Applications 2024. This manuscript contains an extended case study compared with the preprint version. Please see the Appendix.}}

\author{
  Tiankai Xie*,
  Jiaqing Chen*,
  Yaoqing Yang*,
  Caleb Geniesse*, \\
  Ge Shi, 
  Ajinkya Chaudhari,
  John Kevin Cava, 
  Michael W. Mahoney, \\
  Talita Perciano, 
  Gunther H. Weber, and
  Ross Maciejewski
}
\date{}

\graphicspath{{figs/}{figures/}{pictures/}{images/}{./}} 

\usepackage{tabu}                      
\usepackage{booktabs}                  
\usepackage{lipsum}                    
\usepackage{mwe}                       

\usepackage{mathptmx} 

\usepackage{amsmath,amsfonts}
\usepackage{algorithm}
\usepackage{algpseudocode}

\usepackage{bm}
\usepackage{multirow}
\usepackage{booktabs}                  
\usepackage{lipsum}                    
\usepackage{mwe}                       
\begin{document}


\maketitle
\begin{abstract}\looseness-1Modern machine learning often relies on optimizing a neural network's parameters using a loss function to learn complex features. Beyond training, examining the loss function with respect to a network's parameters (i.e., as a loss landscape) can reveal insights into the architecture and learning process. While the local structure of the loss landscape surrounding an individual solution can be characterized using a variety of approaches, the global structure of a loss landscape, which includes potentially many local minima corresponding to different solutions, remains far more difficult to conceptualize and visualize. To address this difficulty, we introduce \textbf{\SYSNAME}\footnote{
Authors with * have equal contributions. \\
Tiankai Xie, Jiaqing Chen, John Kevin Cava, and Ross Maciejewski are with Arizona State University. \\
\emph{E-mail: \{txie21, jchen501, jcava, rmacieje@asu.edu\}@asu.edu}. \\
Yaoqing Yang is with Dartmouth College. \\
\emph{E-mail: yaoqing.yang@dartmouth.edu.} \\
Ge Shi and Ajinkya Chaudhari are with the University of California, Davis. \\ \emph{E-mail: geshi@ucdavis.edu, ajinkyajc@gmail.com.} \\
Caleb Geniesse,  Talita Perciano, and Gunther H. Weber are with Lawrence Berkeley National Laboratory. \\
\emph{E-mail: \{cgeniesse, tperciano, ghweber\}@lbl.gov.} \\
Michael W. Mahoney is with ICSI, LBNL, and the University of California at Berkeley. \\
\emph{E-mail: mmahoney@stat.berkeley.edu.}
}, a visual analytics framework that explores loss landscapes at multiple scales. \textbf{\SYSNAME} integrates metrics from global and local scales into a comprehensive visual representation, enhancing model diagnostics. We demonstrate \textbf{\SYSNAME} through two case studies: visualizing how residual connections influence a ResNet-20, and visualizing how physical parameters influence a physics-informed neural network (PINN) solving a simple convection problem.

\end{abstract}

\section{Introduction}
\label{sec:introduction}

The success of neural network models in natural language processing~\cite{vaswani2017attention} and computer vision~\cite{he2016deep} is often attributed to progressively intricate model architectures and increasing volumes of data. A key step involves training a neural network to learn complex features from data. The objective used in this learning process is known as a \emph{loss function}, written here as $\mathcal{L}_{train}(\theta)$, which quantifies the mismatch between a network's output and the ground-truth (or target) value. The loss function measures how good a model using the set of weights $\theta$ is at predicting the target, and model weights are adjusted during training, such that the loss function is minimized. 

Characterizing the loss function with respect to the weights of a neural network, i.e., as a \emph{loss landscape}, has become an increasingly popular tool for interpreting and understanding the properties of neural network models~\cite{martin2021predicting}.  Many papers have utilized loss landscapes to gain insights into model architecture, training dynamics, and robustness~\cite{li2018visualizing}. However, the construction of loss landscapes and the metrics used to analyze them vary significantly between different approaches and continue to evolve and improve. For example, linear interpolation is used to visualize the loss landscape by creating a randomized subspace~\cite{li2018visualizing}, sharpness is used to measure the quality of the loss landscape~\cite{yao2020pyhessian}, and Hessian spectral analysis is used to study the statistical properties and robustness of trained models~\cite{yao2021adahessian}.
While loss landscape research has proposed many methods and metrics for measuring model properties, and, e.g., loss landscape metrics such as those based on the Hessian can reflect things like the local flatness of the loss landscape, a recent paper~\cite{yang2021taxonomizing} showed that a single property alone cannot explain the full structure of the loss landscape or explain phenomena such as generalizability.
Yang et al.~\cite{yang2021taxonomizing} characterized both the \emph{local} and \emph{global} structure of loss landscapes, and proposed a taxonomy based on metrics at different scales to capture different aspects of the loss landscape. 
They considered both local and global information, where local information refers to the study of specific trained models, e.g., a single SGD solution, and global information describes things about the entire space of possible solutions. They also note that while measuring global properties is more challenging, the combination of (pairwise) similarity and connectivity metrics is sufficient to reveal insights about the global structure of the loss landscape. 
Importantly, while the proposed taxonomy highlights the importance of considering both local and global metrics, interpreting the relationship between these metrics and drawing meaningful conclusions about the model's behavior remains challenging. 
Thus, there is a growing need for a systematic framework specifically designed for multi-scale loss landscape analysis. Such a workflow should provide a cohesive representation of the landscape, integrating model properties across different levels of detail.
By enabling researchers and practitioners to examine their models at multiple scales and from various perspectives, a well-designed visual analytics framework can significantly enhance the interpretability and utility of loss landscapes for model development.

Motivated by this need, we introduce \textbf{\SYSNAME}, a visual analytics framework that provides an interactive, multi-scale visual representation and analytical pipeline for exploring the loss landscapes of machine learning models. Building on the taxonomy of loss landscapes proposed by Yang et al., \textbf{\SYSNAME} is designed to make it easier for machine learning researchers and practitioners to examine their models across two distinct levels: \emph{local} and \emph{global} scale.
At the local scale, \textbf{\SYSNAME} provides single-model-level metrics and localized views of the loss landscape surrounding an individual trained model. This allows users to analyze the sharpness, curvature, and other properties of the landscape in the vicinity of a specific solution. At the global scale, the system attempts to summarize the global landscape as a fully connected graph, combining all trained models and using a two-dimensional layout based on Centered Kernel Alignment (CKA) similarity. This global view enables users to identify clusters of similar models, explore the relationships between different solutions, and gain a broader understanding of the optimization process.
\textbf{\SYSNAME} allows experts to load any number of trained models for a specified model configuration (i.e., same architecture and dataset), as well as models for multiple different configurations simultaneously. By providing a cohesive visual representation that links metrics evaluated at different scales, the system helps users form connections between local and global properties of the loss landscape and extract novel insights about different aspects of model development.

\begin{figure}[t!]
    \centering
    \includegraphics[width=0.60\columnwidth]{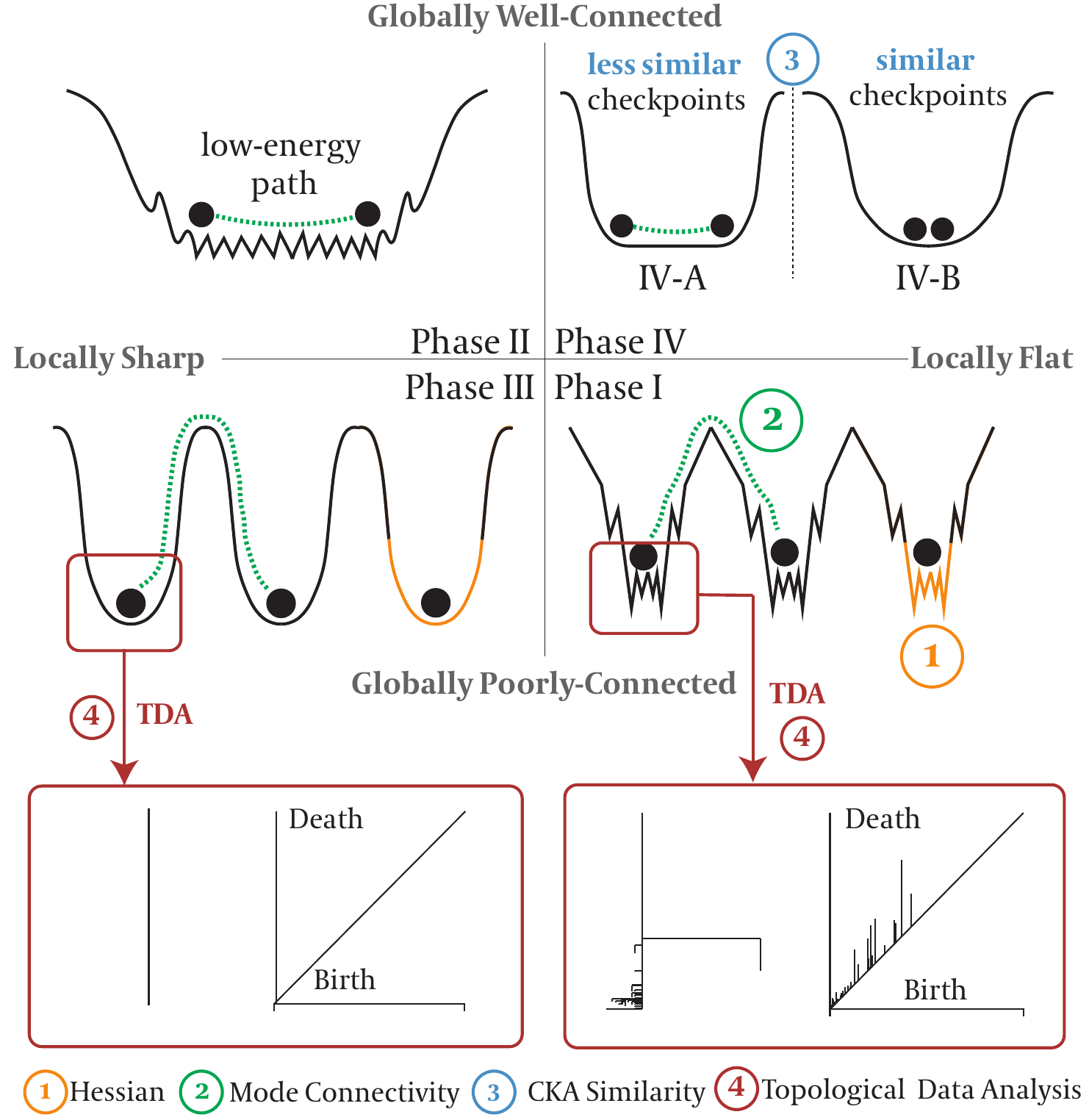}
    \caption{Caricature of different types of loss landscapes. 
    Loss landscapes can be classified into five types~\cite{yang2021taxonomizing}: globally well-connected versus globally poorly-connected loss landscapes; and locally sharp versus locally flat loss landscapes, where locally flat and globally well-connected loss landscapes can be further distinguished based on the similarity between models.
    Globally poorly connected loss landscapes have high barriers between minimum loss areas, and globally well-connected loss landscapes tend to have low-energy (low-loss) paths between models. Locally flat loss landscapes are ``smoother'' than locally sharp landscapes. Metrics that are used for categorizing the types are \textcircled{\raisebox{-.9pt}{1}} Hessian; \textcircled{\raisebox{-.9pt}{2}} Mode Connectivity; \textcircled{\raisebox{-.9pt}{3}} CKA Similarity and \textcircled{\raisebox{-.9pt}{4}} Topological Data Analysis. Here, local sharpness is measured by the Merge Tree (left) and the Persistence Diagram (right). Multiple branches of the merge tree and rough persistence diagram indicate the ``bad'' landscape, and vice versa.
    }
\label{fig:loss-landscape-types}
\end{figure} 
Our contributions include: 
\begin{enumerate}
    \item 
    A visual analytics framework, \textbf{\SYSNAME}, for diagnosing models by exploring multi-scale loss landscapes across local and global scales.
    \item 
    A cohesive visual representation that links metrics evaluated at these different scales, forms connections between metrics for global and local loss landscapes.
    \item
    Two case studies that demonstrate how \textbf{\SYSNAME} can be used to extract novel insights from neural networks' loss landscapes about different aspects of model development. 
\end{enumerate}

\section{Background}
\label{sec:background}

\vspace{1mm}

In this section, we discuss key machine learning concepts and techniques used as components in \textbf{\SYSNAME}. 
We also review concepts from topological data analysis that we use to analyze loss functions.

\subsection{Machine Learning Concepts}
\label{sec:background_loss}
\vspace{1mm}

The general \emph{loss function} of a neural network is:
\begin{equation}
  \label{eq:loss_function}
    \vspace{-1mm}
  \mathcal{L}(\theta) = L(D, \theta) =  \frac{1}{n} \sum^{n}_{i=1} l(x_i,y_i;\theta),
  \vspace{-1mm}
\end{equation}
where ${x_i, y_i} \in D$ represents a sample in the training dataset $D$, with $x_i$ as the feature vector and $y_i$ as the corresponding label. $\theta$ denotes the neural network's parameter weights, and $n$ is the number of samples.

\vspace{1mm}
\noindent 
The \emph{Hessian} is the second-order derivative of the loss function, $\nabla^2_\theta \mathcal{L}(\theta)$, with respect to weights $\theta$.
The Hessian captures the local curvature of the loss landscape~\cite{yang2021taxonomizing}, and it can be used to differentiate locally sharp versus locally flat loss landscapes as shown in Figure~\ref{fig:loss-landscape-types}.1.
The eigenvalues of the Hessian are the principal curvatures of the loss function. In this work, we capture the local curvature of the loss landscape by calculating the top Hessian eigenvalues, computed as $\{\lambda_i(\nabla^2_\theta \mathcal{L}(\theta)), i \in [1, 10]\}$, using \texttt{PyHessian}~\cite{yao2020pyhessian}.

\vspace{1mm}
\noindent \emph{Mode Connectivity}~\cite{garipov2018loss}
refers to the existence of curves or pathways that connect the optima (modes) of loss landscapes in a neural network.
These curves can be represented by simple geometric shapes, such as polygonal chains with only one bend.
Mode Connectivity provides a new understanding of the geometric properties of neural network loss surfaces, and it has significant implications for training efficiency, ensembling techniques, and Bayesian deep learning~\cite{garipov2018loss}.
Computing the mode connectivity between two sets of weights $\theta$ and $\theta'$ involves finding a low-loss curve $\gamma(t), t \in [0, 1]$, where $\gamma(0) = \theta$ and $\gamma(1) = \theta'$, such that $\int\mathcal{L}(\gamma(t))dt$ is minimized.
Given the curve represented as $\gamma_{\phi}(t)$, we define the mode connectivity between $\theta$ and $\theta'$ to be
\begin{align}\label{eq:MC}
      mc(\theta,\theta') = \frac{1}{2}(\mathcal{L}(\theta) + \mathcal{L}(\theta')) - \mathcal{L}(\gamma_{\phi}(t^{*})),
\end{align}
where $t^{*}$ maximizes the deviation $ t \mapsto |\frac{1}{2}(\mathcal{L}(\theta) + \mathcal{L}(\theta')) - \mathcal{L}(\gamma_{\phi}(t))|$. 
In general, $mc(\theta, \theta') < 0$ indicates the existence of a high loss ``barrier'' between $\theta$ and $\theta'$, implying the mode connectivity is \emph{poor} and the loss landscape is considered to be poorly-connected. $mc(\theta, \theta') > 0$ indicates the existence of a (relatively) lower loss path between $\theta$ and $\theta'$, which implies that the initial training failed to find reasonable optima. $mc(\theta, \theta') \approx 0$ indicates a \emph{good} mode connectivity. In this case, we say the loss landscape is ideally well-connected.
In Figure~\ref{fig:loss-landscape-types}.\raisebox{.5pt}{\textcircled{\raisebox{-.9pt}{2}}}, we show globally poorly-connected loss landscapes in the top row and globally well-connected loss landscapes in the bottom row. Globally poorly-connected loss landscapes have high barriers between different local minima, which can be measured using mode connectivity.

\vspace{1mm}
\noindent \emph{CKA Similarity}~\cite{kornblith2019similarity} is a metric
for evaluating the similarity between two sets of features learned by two neural networks.
CKA similarity is widely used because it is invariant to orthogonal transforms and isotropic scaling, which is a desired property to deal with rotation and scaling of feature representations.
It is more robust to variations in network architectures and hyperparameters, compared to other similarity metrics such as cosine similarity or correlation coefficient~\cite{kornblith2019similarity}.
Formally, for a neural network $f_\theta$ with weights $\theta$, let $F_\theta = \begin{bmatrix}f_\theta(\boldsymbol{x}_1)&\cdots &f_\theta(\boldsymbol{x}_m)\end{bmatrix}^\top \in \mathbb{R}^{m\times d}$ denote the concatenation of the (vectorized) feature maps of length-$d$ of the network over a set of $m$ randomly sampled data points.
Then the (linear) CKA similarity between two sets of feature maps is given by
\begin{align}\label{eq:CKA}
    s(F_\theta,F_{\theta'}) = \frac{\text{Cov}(F_\theta, F_{\theta'})}{\sqrt{\text{Cov}(F_\theta, F_{\theta})\text{Cov}(F_{\theta'}, F_{\theta'})}},
\end{align}
where we define $\text{Cov}(X,Y) = (m-1)^{-2}\text{tr}(XX^\top H_m YY^\top H_m)$, and $H_m = I_m - m^{-1}\mathbf{1}\mathbf{1}^\top$ is the centering matrix, for $X,Y\in \mathbb{R}^{m\times d}$. 
We note that the feature maps $F_\theta$ and $F_{\theta'}$ can be taken from any layer of two neural networks. They can even be taken from two layers of the same neural network.
When comparing two neural networks $\theta$ and $\theta'$, we often represent the CKA similarity using a layer-wise similarity matrix, where the ($i$,$j$)-th coordinate represents the similarity between the feature maps taken from the $i$-th layer of $\theta$ and the  $j$-th layer of $\theta'$. 
In the upper-right corner of Figure~\ref{fig:loss-landscape-types}, 
CKA similarity further divides the globally well-connected and locally flat loss landscapes into two subtypes.

\subsection{Topological Data Analysis}
\vspace{1mm}

Topological data analysis (TDA) offers insights into the overall patterns of functions and relies on the fundamental concept of ``connectedness.''
In the context of the loss function, we are interested in how many optima/minima exist in the loss function and how ``prominent'' these minima are, measured by their persistence (defined below).
This information is captured by $0$-dimensional persistent homology and the so-called merge tree.
The \emph{Merge Tree}~\cite{heine2016survey} tracks connected components of \emph{sub-level sets} $L^-(v) = \{ x \in D; x \le v\}$.
As the threshold $v$ increases, new connected components form at minima and later merge with neighboring connected components at saddles.
The merge tree records these changes, representing minima where newly connected components form as degree-one nodes and merges of two connected as degree-three nodes.
See Figure~\ref{fig:loss-landscape-types}.\raisebox{.5pt}{\textcircled{\raisebox{-.9pt}{4}}}.

Because of their connection to persistent homology~\cite{edelsbrunner2008persistent},
it is possible to derive the $0$-dimensional persistence diagram from the branches of the merge tree.
The persistence diagram represents branches, i.e., features, as points in a two-dimensional plane. 
The horizontal axis represents the birth function value of the feature, which is the value of the minimum at which it first appears. 
The vertical axis represents the death function value of the feature, which is the value of the saddle where it merges into a more persistent feature.
The distance between a point and the diagonal line $y = x$ represents the persistence of the feature, which is a measure of how long it lasts in the data. 

Our paper aims to provide the first connection between TDA methods and those used by Yang et al.~\cite{yang2021taxonomizing} in studying and taxonomizing loss landscape structures. 
As shown in Figure~\ref{fig:loss-landscape-types}, both merge trees and persistence diagrams can provide complementing information on the ruggedness of a loss landscape, in addition to methods such as mode connectivity and CKA similarity. 
For instance, persistence diagrams can show not only the multiple scales of loss minima but also the relative scales between loss minima that contain each other, which visualizes the proportions of deep versus shallow loss minima and the comparison between them.
In addition, merge trees can complement the loss landscape information by characterizing the multi-scale structure and the connected patterns between these loss minima using a tree representation, which significantly extends CKA similarity and mode connectivity.

\section{Related Work}
\label{sec:related_work}

\subsection{Visualizing and Measuring Loss Landscape}
\label{sec:relate_work_loss_landscape}

\vspace{1mm}

Visualizing high-dimensional loss functions of neural networks in a human-perceptible way is challenging.
Goodfellow et al.~\cite{goodfellow2014qualitatively} adopted a one-dimensional linear interpolation method, plotting the loss surface between a randomly initialized model and a nearby minimizer obtained via stochastic gradient descent. 
This method has been extensively used to explore various aspects of the loss function, such as the sharpness and flatness of different minima~\cite{dinh2017sharp}, the relationship between sharpness and batch size~\cite{keskar2016large}, the peaks between different minima~\cite{smith2017exploring} and minima obtained via different optimizers~\cite{im2016empirical}.
Goodfellow et al.~\cite{goodfellow2014qualitatively} also proposed a two-dimensional visualization approach, mapping loss values onto a plane spanned by two random directions in the weight space.
Li et al.~\cite{li2018visualizing} improved the resolution of loss landscapes with filter-wise normalization, which removes scaling effects.

Yao et al.~\cite{yao2020pyhessian} conducted a large-scale empirical study on Hessian-based methods for analyzing learning models.
To visualize the largest loss variations in the weight space, they project the model to the two-dimensional space spanned by the two top eigenvectors of the Hessian.
Quantitatively measuring the structure of loss landscapes can sometimes be more helpful, and sharpness and flatness are frequently used to study trained model quality and enhance performance~\cite{keskar2016large,yao2018hessian}. It is often believed that when a model converges to a sharp local minimum, it results in a large gap between training and test accuracies, leading to poor performance on unseen test data.
Recent work by Doknic et al.~\cite{doknic2022funnscope} presents FuNNscope, a tool for interactively exploring the loss landscape of fully connected neural networks by visualizing the loss landscape through axis-parallel slicing. However, their definition of loss landscape remains on a local scale, without addressing the global scope.

In contrast, our visual analytics system uses both the local and global structure of loss landscapes. The addition of the global component is critical as previous work suggests that global structural information can provide more useful guidance compared to local alone.

\section{Design Overview}
\label{sec:system_overview}
\vspace{1mm}

In this section, we outline our system design.
We begin by highlighting the significance of investigating the loss landscape through global and local analyses.
This is achieved by summarizing several essential analytical tasks frequently employed by machine learning practitioners to measure the structural information of a loss landscape.
Subsequently, for each set of analytical tasks, we refined a set of design requirements and system components for our framework alongside our ML collaborators to address these analytical tasks.

\subsection{Analytical Tasks}
\vspace{1mm}

Generally speaking, a smoother loss landscape (similar to a well-conditioned quadratic curve) makes for easier optimization, while a rougher loss landscape (with abundant spurious local minima and sharp curvature directions) makes optimization more difficult~\cite{li2018visualizing}.
Using this intuition, we identified analytical tasks to obtain critical loss landscape information for model diagnostics.
However, being smooth or rough can have different meanings, depending on the scale.

\vspace{1mm}
\noindent \textbf{T1. Global Analysis of the Loss Landscape.}
Given a relatively global picture measured on many local minima, obtained by stochastic optimization with diverse configurations of random seeds or training hyperparameters, analysts explore the following questions:
\begin{itemize}
\item 
\textbf{T1.1} Are these local minima well-connected by low-loss pathways?
\item 
\textbf{T1.2} When models are trained with full data, will different models converge to ``essentially the same'' function (in terms of mapping a fixed input sample to similar outputs)?
\end{itemize}
The first analytical task characterizes the roughness on a relatively global scale.
For example, if the loss minima are all poorly connected to each other, a machine learning practitioner may conclude that stochastic optimization can be stuck at a spurious low-quality local minimum.
We use mode connectivity to differentiate between well-connected and poorly-connected local minima in this scale.
The second task asks if the training data is adequate to obtain a sufficiently optimized learning model or if the data is too scarce, leading to ``underspecified'' learning models.
Insufficient training data could lead to a smaller CKA similarity between trained models~\cite{yang2021taxonomizing}, and we use this metric for T1.2.

\vspace{1mm}
\noindent \textbf{T2. Local Analysis of the Loss Landscape.}
Given a local picture, measured around a single model, analysts explore the following:
\begin{itemize}
    \item 
    \textbf{T2.1} Is the minimum around the model flat (or sharp), as measured using Hessian?
    \item 
    \textbf{T2.2} Is the minimum smooth in the loss landscape visualization?
\end{itemize}

\begin{figure*}[tbh]
    \centering	
    \includegraphics[width=\linewidth]{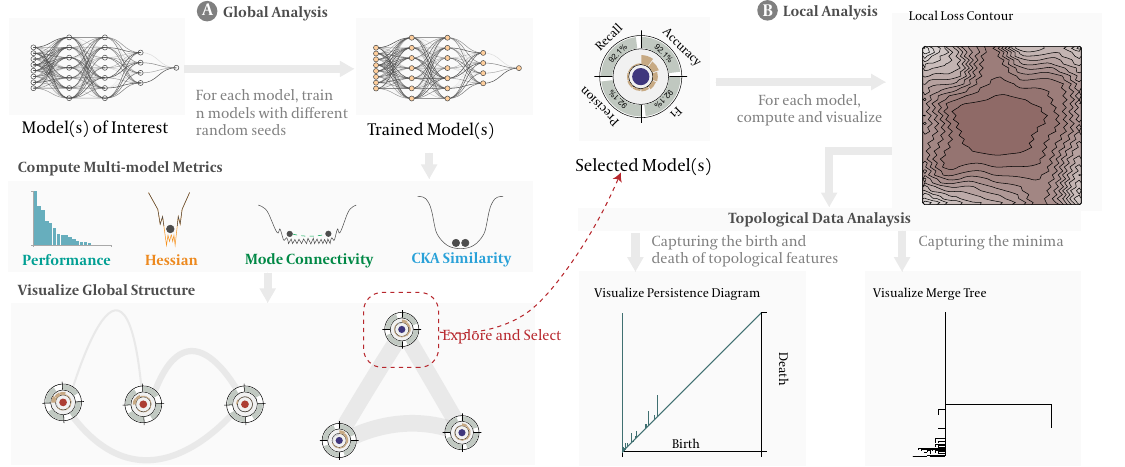}
    \vspace{-2mm}
    \caption{Our framework consists of two stages: 
    (A) Global Analysis, and
    (B) Local Analysis.
    In Stage A, analysts evaluate a model's global structure by computing a combination of metrics for trained models and exploring the visualized global structure for models of interest. 
    Next, in Stage B, analysts investigate the detailed local features of a selected model's loss landscape. 
    To mitigate the situation when visualized loss contours are hard to evaluate, persistence diagram and merge tree visualization techniques are applied to help analysts better analyze the local structure.
    \label{fig:visual_analytics_framework}
    }
\vspace{-4mm}
\end{figure*}

\subsection{Design Requirements}
\vspace{1mm}

We worked closely with domain experts to iterate through various designs. Through discussions, prototyping, and feedback, we linked these analytical tasks to a specific set of design requirements.

\vspace{1mm}
\noindent \textbf{D1. Visualize the global picture of the Loss Landscape.}
We visualize the global structure by exploring a scale of loss minima, obtained by different random seeds for weight initialization and optimization stochasticity. We then combine mode connectivity (\textbf{T1.1}) and CKA similarity (\textbf{T1.2}) by representing them as a graph that we call \emph{Global Structure View}, see Figure~\ref{fig:resnet_case_study}.A.

\vspace{1mm}
\noindent \textbf{D2. Visualize the local picture of the Loss Landscape.}
When we restrict ourselves to a local picture, we can precisely define the flatness of a loss minimum using the Hessian (\textbf{T2.1}), as Hessian is often used for assessing the generalizability of learned models~\cite{yao2018hessian,yao2021adahessian}. 
However, the Hessian is measured on a single high-dimensional point, and using Hessian information alone can make the local analysis overly restricted. 
This motivates us to visualize the \emph{roughness} of the loss landscape around a single model, measured by the loss contour and the TDA analysis. 
We visualize the local structure using the following components (\textbf{T2.2}):
\begin{itemize}
    \item 
    The \textit{Loss Contour} plot presents the loss landscape, i.e., a two-dimensional slice of the high-dimensional loss function. 
    \item 
    The \textit{Persistence Diagram} shows the persistent homology of the loss landscape~\cite{cohen2005stability}. 
    \item 
    The \textit{Merge Tree} summarizes topological structure of the loss landscape~\cite{heine2016survey}.
\end{itemize}

\section{\SYSNAME~ Framework}

\label{sec:visual-analytics-framework}
\vspace{1mm}

\textbf{\SYSNAME}, depicted in Figure~\ref{fig:visual_analytics_framework}, consists of two aspects: (A) Global Analysis and (B) Local Analysis.
Analysts upload models and explore the global structure of the loss landscape constructed by sampled models. They investigate model(s) of interest in detail by visualizing the local structure of the loss landscape(s) and performing comparisons between selected models in terms of their architecture and predictions. 

\subsection{Global Analysis}
\label{sec:augmentation_overview}

Analysts are primarily interested in effectively gaining insights from the global loss landscape structure. 
The most common approach to achieve this objective is visualizing the local structure,
meaning the graphical representation of how a particular loss function changes with respect to the model's parameters.
Our approach enables the exploration of local versus global loss landscape structure. Three key metrics (defined in the Background Section) are involved: Hessian for measuring local sharpness; mode connectivity for evaluating the connectivity between two minima, and; CKA similarity for evaluating the disparity in predictions.

\begin{figure}[t!]
    \centering
\includegraphics[width=0.60\columnwidth]{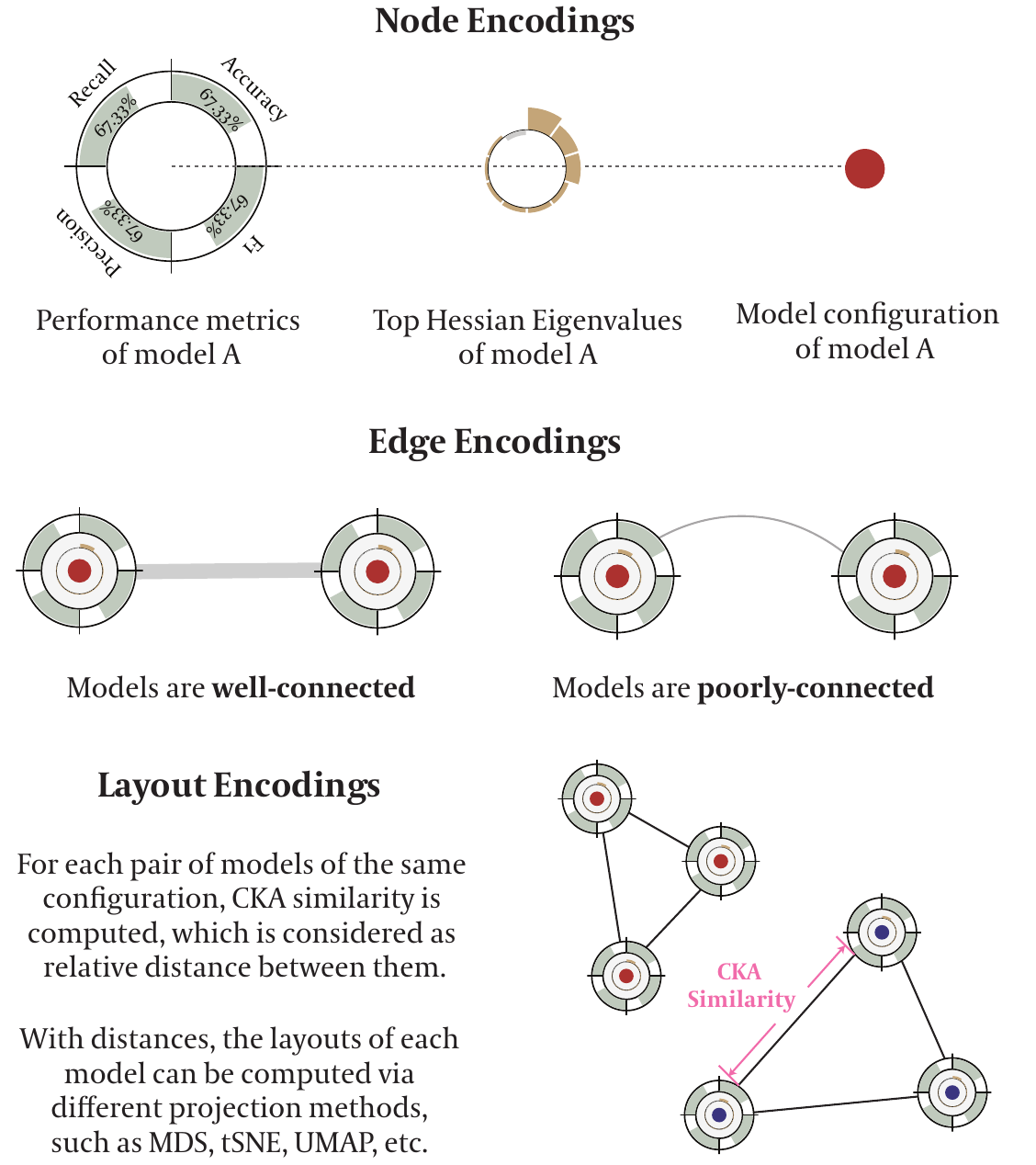}
    \caption{Visual encodings in the global structure view. 
    Each node in the global structure view has three layers: The outer ring visualizes the performance of the model, e.g., accuracy, recall, precision, and F1; the middle ring visualizes the top Hessian eigenvalues of the model; and the color of the inner circle represents one unique model so that analysts can distinguish between different models of interest. Edge properties represent the mode connectivity between models: a thick and straight line implies good mode connectivity and a thin and curved line implies poor mode connectivity. 
    The position of each model in the global structure view is visualized by projection methods along with the CKA similarity.
    }
\label{fig:global_structure}
\vspace{-6mm}
\end{figure}

\vspace{0.5mm} 
\noindent \textbf{Global Structure View.}
We design a global structure view to effectively convey these metrics. As shown in Figure~\ref{fig:visual_analytics_framework}.A, we first train $n$ models given the tasks of interest, where $n$ can be any number based on analysts' preferences. 
Analysts can generate many models for each model configuration. The more models the analysts sample, the more concrete the model's global structure is. Next, we use sampled models to construct the global structure considering \emph{multi-model}, \emph{two-model}, and \emph{single-model} metrics (Figure~\ref{fig:global_structure}). 

\vspace{0.5mm} 
\noindent \emph{Multi-model Metric.}
For the global metric, we compute the CKA similarity as the relative distance for each pair of models of the same model configuration, and a variety of projection methods can be applied to generate the layout of the view, such as multi-dimensional scaling (MDS)
,
t-Distributed Stochastic Neighbor Embedding (tSNE)
, and Uniform Manifold Approximation and Projection (UMAP)
. Since CKA similarity is architecture agnostic, the metric also applies to the similarity between the features from different architectures (\textbf{T1.2}).

\noindent \emph{Two-model Metric.}
We compute the mode connectivity as the two-model metric to obtain the connectivity between a pair of models. Since the mode connectivity tries to find the low-energy path between two models, we compute the mode connectivity between all models within the same model configuration, which yields a numeric value for each pair of models. When this number is larger, the models are better connected. We use both thickness and degree of curvature of connecting lines (Figure~\ref{fig:global_structure}) to represent the mode connectivity: When two models are well-connected, the edge between them is thicker and straighter; otherwise, the edge gets thinner and more curved (\textbf{T1.1}).

\noindent \emph{Single-model Metric.}
We also consider integrating a variety of local metrics for each model to facilitate analysts to obtain overall local properties (\textbf{T2.1}). As shown in Figure~\ref{fig:global_structure}, each model is encoded as a circle with three levels of information: we integrate performance metrics of the model to the outer ring of the circle, such as test accuracy, recall, precision, and f1 score. The performance metrics can be customized based on the task type of the model. If the model is for a regression task, the performance metrics can be MSE, MAE, etc. We then integrate the $k$ largest Hessian eigenvalues of the models to reveal their local loss landscape curvature. The number $k$ can also be customized. In this work, we obtain the top-$10$ Hessian eigenvalues for demonstration purposes.
Finally, the color represents the different models. Our framework supports providing global structure across multiple models with and without the residual links. Analysts can add an arbitrary number of models for global structure comparison.

\begin{figure}[t!]
    \centering
\includegraphics[width=0.60\columnwidth]{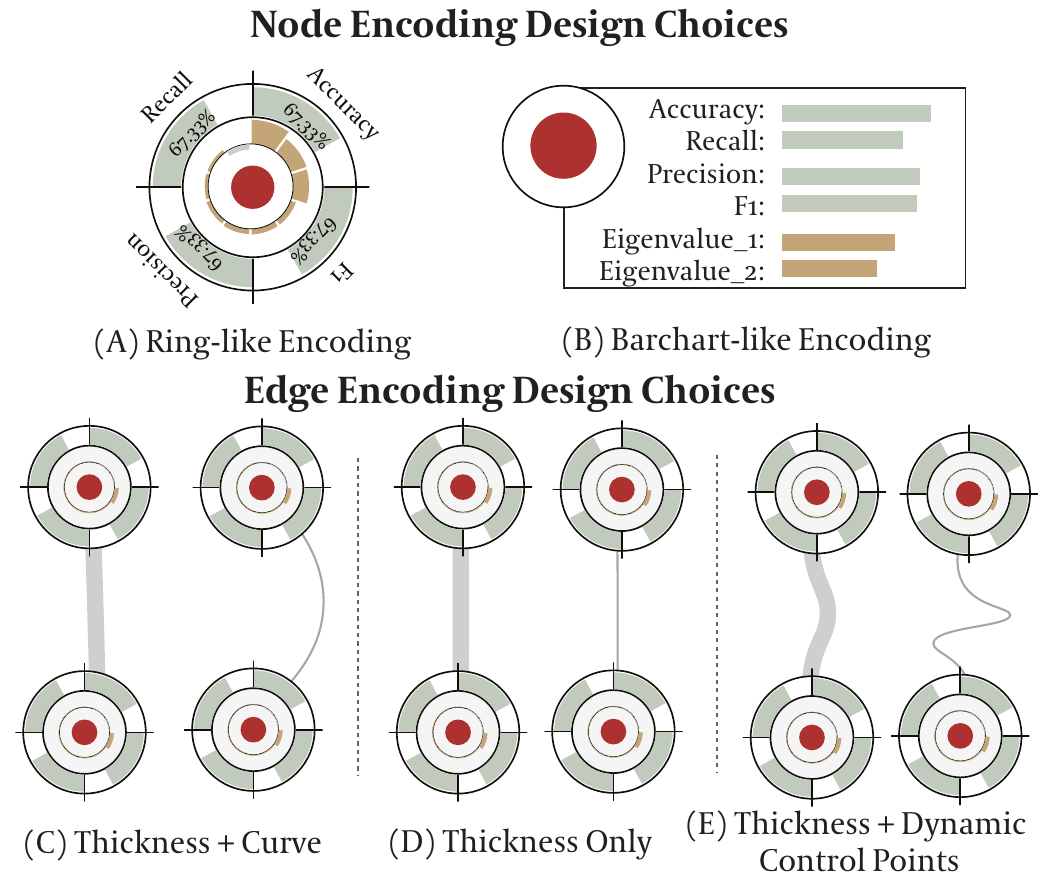}
    \caption{Global Structure Design: Versions A and C are preferred by experts.
}
    \label{fig:global-structure-design-choices}
\end{figure}

\noindent \textbf{Alternative Design.}
We have also considered several alternative designs for the global structure view (Figure~\ref{fig:global-structure-design-choices}). We tried encoding all one-model metrics as a list of bar charts (Figure~\ref{fig:global-structure-design-choices}. B), while all the metrics are comparable and simple to understand, the graph can create occlusion. The ring-like encoding (Figure~\ref{fig:global-structure-design-choices}. A) was better for the domain experts to see those bars as local metrics and labels next to the bars were added for fast recognition per our experts' request Figure~\ref{fig:global-structure-design-choices}. According to our collaborating domain experts, this approach better represents the global structure of the loss landscape. In their research, this visualization can help them clearly identify the important concepts. For example, model performance and the top Hessian eigenvalues are displayed around the node since they describe the local properties of that individual model, mode connectivity is displayed as a path between two models since it reflects how well they are connected, and the relative distance between models represents the pairwise similarity between all of the models. Thus, the global structure representation can provide comprehensive information about the loss landscape in a meaningful way. 

\noindent \textbf{Interactions.}
The design effectively represents key properties of the loss landscape, but such visualization can lead to issues like visual overlap. We provide various interactions to address scalability when sampling more trained models. We have toggles that hide or show the different types of information in the global structure view: (i) \emph{performance labels}  controls the accuracy labels outside the circle; (ii) \emph{performance information} controls the outer ring representing model performance; and (iii) \emph{Hessian information} controls the middle ring showing local loss landscape curvature. Hiding information helps to avoid potential circle overlap. Our global structure view also supports zooming and panning focus on individual circles clearly. Most visualization components have tooltips showing detailed information like accuracy, Hessian eigenvalues, and mode connectivity.  

\subsection{Local Analysis}
We apply three methods to visualize the local loss landscape (\textbf{T2.2}).
First, a \emph{loss contour view} enables analysts to obtain an overall picture of the loss landscape around a single model.
Second, a \emph{persistence diagram} captures the loss contour's topological features and their evolution across different spatial scales or parameter values when different loss contours are visually hard to compare, as the persistence diagram provides a concise and informative summary of the homological features.
Third, a \emph{merge tree} view captures the hierarchical structure of the loss contour when the persistence diagram is too visually subtle to distinguish. 
While Yang et al.~\cite{yang2021taxonomizing} utilize the Hessian as the local curvature metric of the loss landscape, we incorporate the previous loss landscape generating method~\cite{goodfellow2014qualitatively, li2018visualizing} to provide the loss contour view from a different angle. 

\begin{figure*}[ht!]
\centering
    \includegraphics[width=\linewidth]{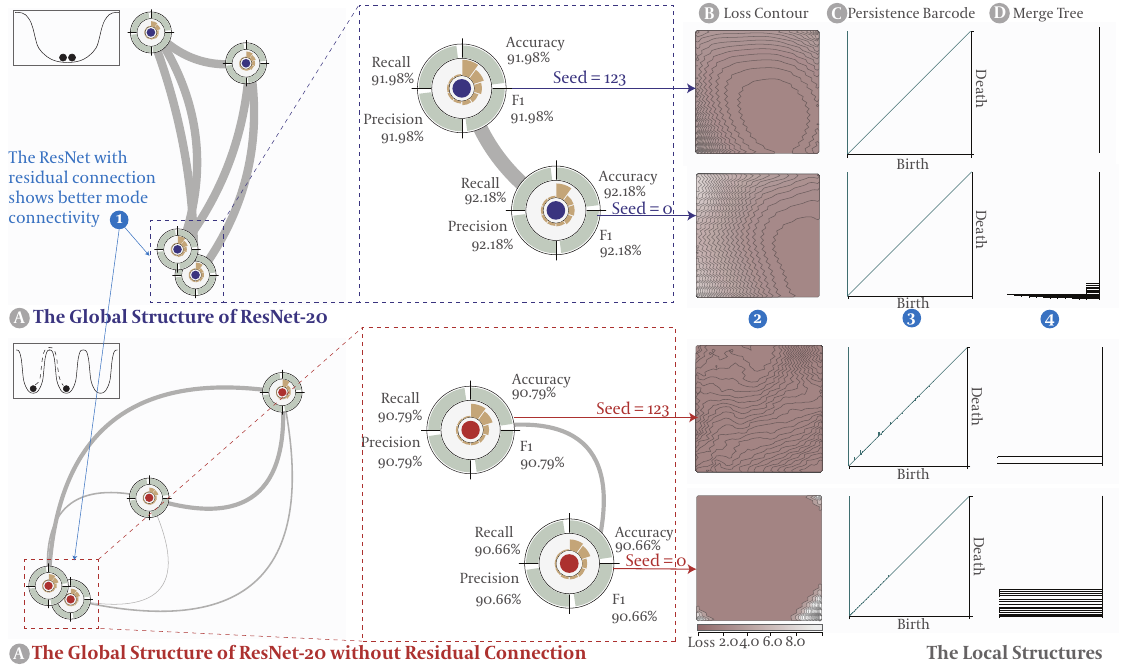}
    \vspace{-2mm}
    \caption{
   \textbf{\SYSNAME} compares ResNet-20 with and without residual connections, {\textcircled{\raisebox{-.9pt}{1}}} revealing that despite similar accuracy (90-92\%), ResNet-20 with residuals (ResNet-20-R) exhibits better mode connectivity,{\textcircled{\raisebox{-.9pt}{2}}} a flatter loss landscape, and {\textcircled{\raisebox{-.9pt}{3}}} {\textcircled{\raisebox{-.9pt}{4}}} more branches in the merge tree compared to ResNet-20 without residuals (ResNet-20-NR). This highlights how residual connections improve the ResNet architecture beyond performance.}
    
    \label{fig:resnet_case_study}
\end{figure*}

\vspace{0.5mm} 
\noindent \textbf{Loss Contour.} 
In the machine learning community, the most common means of visualizing the loss contour is by projecting the loss to a one-dimensional or two-dimensional space~\cite{li2018visualizing}.%
In this paper, we visualize the loss contour in a two-dimensional projection. The model parameters are mapped onto two random vectors, and each projection is subdivided into 20 steps where each element of the array denotes a loss value. 
We use dark brown to denote lower loss and light brown to denote higher loss.
When the model contains BatchNorm layers, e.g., Figure~\ref{fig:resnet_case_study}.B, naively plotting the loss contour generates numerical errors because the running statistics of the learned model do not apply to its projections on the hyperplane.
To address this issue, we adopt a warm-up phase before the actual loss evaluation to update and correct running statistics in the BatchNorm layer.
For each element of the projected model, we call its forward pass by setting the running statistics as trainable while fixing the weights.

\vspace{0.5mm} 
\noindent \textbf{Alternative Design.}
Our goal is to visualize the loss contour in a way the domain experts are familiar with. As such, we adopt the method from Goodfellow et al.~\cite{goodfellow2014qualitatively} to plot the loss landscape. We have also tried visualizing the three-dimensional loss contour as the alternative design, where loss values of the landscape are encoded as the height along the $z$-axis. Although such a design shows a concrete representation of the loss landscape, it becomes hard to perceptualize when the landscape gets complicated, and our collaborators indicated a preference for the two-dimensional format.
\vspace{0.5mm}

\noindent \textbf{Persistence Diagram and Merge Tree.}
Since loss contour can be perceptually hard to understand and compare, \textbf{\SYSNAME} uses TDA to help overcome such challenges.
Using the persistence diagram (Figure~\ref{fig:resnet_case_study}.C), analysts can track each loss minimum's lifetime through the persistence diagram, measuring the loss landscape's topological information, such as peaks, plateaus, rough terrain, etc.
We also use the merge tree to conduct a hierarchical analysis of the loss landscape (Figure~\ref{fig:resnet_case_study}.D).
In this work, both the persistence diagram and merge tree are generated using the two-dimensional slice of the loss landscape. Future iterations will focus on higher-dimensional projections to reduce information loss.

\vspace{1mm}
\noindent \textbf{Interactions.}
For local analysis such as loss contour, persistence diagram and merge tree, we allow analysts to click on the nodes in the global structure view. The merge tree and persistence diagram also support zooming and panning for detailed inspection.

\section{Case Studies}
\label{sec:case_study_and_expert_interview}
\vspace{1mm}

We present two case studies to show how \textbf{\SYSNAME}~can help analysts diagnose machine learning models. As a model-agnostic tool, \textbf{\SYSNAME} supports different diagnostic scenarios. Our case studies are designed to not only show the compatibility of types of tasks but also address the state-of-the-art research questions through the diagnostic process. Overall, our case studies cover two major modeling approaches: 
\emph{architecture alteration}, and 
\emph{loss function alteration}. 
We first investigate the effect of alterations of ResNet-20 with or without residual connection (architecture alteration).
Then, we diagnose how optimizing the loss function can lead to different loss landscapes for PINNs (loss function alteration), which was identified as a reason for PINN failure modes~\cite{krishnapriyan2021characterizing}.

\subsection{Architecture Alteration of ResNet-20}
\label{sxn:resnet}

In our second case study, we utilize \textbf{\SYSNAME} to investigate how architecture can influence the model. 
As Yao et al.~\cite{yao2020pyhessian} point out, the incorporation of neural network components such as batch normalization layers may contradict the widely held belief that the model with such layers will yield a flat loss landscape. Their findings indicate that batch normalization improves model performance when the model is deeper but can hurt performance when the model is shallower. We argue that both observations are correct but underspecified. The outcome depends significantly on the specific parameters and conditions under which the model operates, leading to different phases of behavior. Our objective is to facilitate a clearer visualization of the dynamics at play in these varying scenarios, thereby enabling more precise adjustments to the model parameters. 
In this case study, we conduct further exploration based on Yao et al.~\cite{yao2018hessian} on why residual connections in ResNet-20~\cite{he2016deep} can cause significant changes in the loss landscape. The ResNet-20 model is trained on the CIFAR10 dataset. We denote ResNet-20 with residual connection as ``ResNet-20-R'', and ResNet-20 without residual connection as ``ResNet-20-NR''. We select four arbitrary random seeds (\texttt{0}, \texttt{123}, \texttt{123456}, and \texttt{2023}) to train the models, such that each model has four models based on those seeds.

\begin{figure*}[ht!]
\centering
    \includegraphics[width=\linewidth]{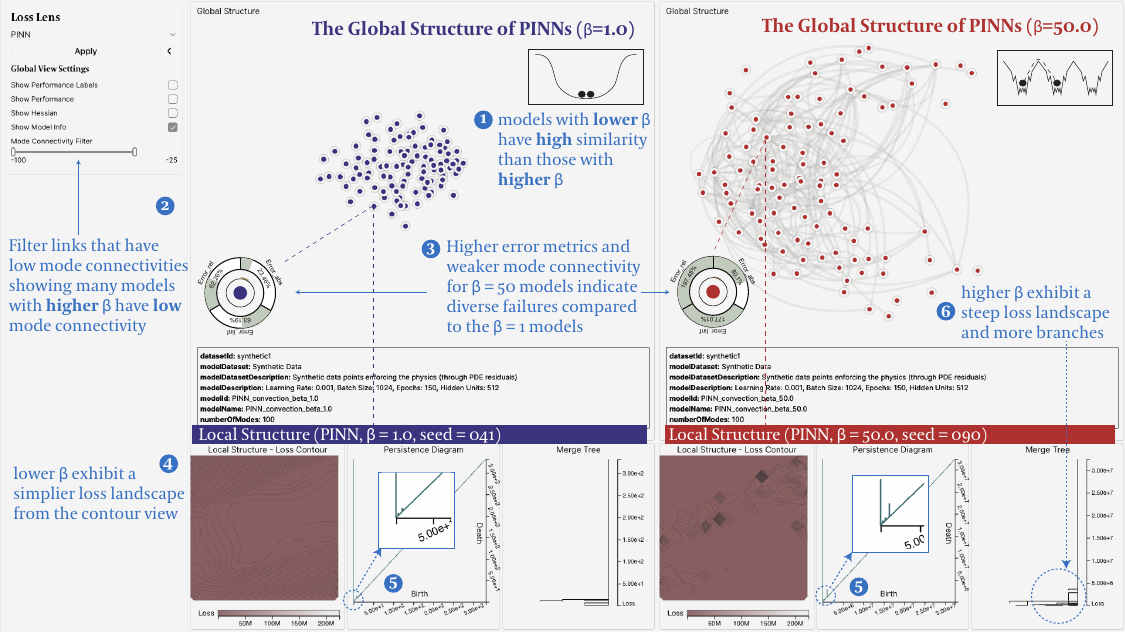}
    \vspace{-2mm}
    \caption{
    We use \textbf{\SYSNAME} to study PINNs solving 1D convection problems with varying wave speeds ($\beta$). For $\beta=50$:
{\textcircled{\raisebox{-.9pt}{1}}} Models are more spread out, indicating diverse failure modes.
{\textcircled{\raisebox{-.9pt}{2}}} Weaker mode connectivity suggests less connected minima.
{\textcircled{\raisebox{-.9pt}{3}}}Higher error metrics confirm solution failure.
{\textcircled{\raisebox{-.9pt}{4}}}  {\textcircled{\raisebox{-.9pt}{5}}} The loss landscape is rugged with sharp peaks of higher loss.
{\textcircled{\raisebox{-.9pt}{6}}} Merge tree shows more branches, indicating increased complexity.
In contrast, $\beta=1$ models show closer proximity, flatter loss landscape, and fewer branches, suggesting a more constrained solution space.
    }
    \label{fig:pinn_case_study}
	
\end{figure*}

\vspace{0.5mm} \noindent \textbf{Global Analysis}.
First, analysts notice that although the average performance of the two models is similar (with an accuracy of approximately 90\% to 92\%), the mode connectivity of ResNet-20-R is better than that of ResNet-20-NR as the lines among the models without residual connection are more curved and thinner (Figure~\ref{fig:resnet_case_study}.1, \textbf{T1.1}).
The analysts wonder if removing the residual connections can significantly affect the loss landscape's smoothness, matching the intuition in prior work~\cite{li2018visualizing}.
For the CKA similarity, analysts noticed that the structure did not show an obvious pattern (Figure~\ref{fig:resnet_case_study}.1).
This observation may imply that \emph{both ResNet-20-R and ResNet-20-NR are trained with a similar amount of data, and the similarity in predictions, measured by the distance between the circles on the global structure view, does not differ significantly (\textbf{T1.2}).}
The observation of the global structures indicates that \emph{ResNet-20-R has a locally flat, globally well-connected landscape, and some models are similar (Figure~\ref{fig:loss-landscape-types}. IV-A)}, while \emph{ResNet-20-NR has a locally flat, globally poorly-connected landscape (Figure~\ref{fig:loss-landscape-types}. III).}

\vspace{0.5mm} \noindent \textbf{Local Analysis}.
It is noteworthy that the models (ResNet-20-NR \texttt{123} and \texttt{0}) are very close to each other. Yet, their connectivity is relatively poor, which indicates the ``high barrier'' between them caused by removing the residual links, while such phenomena do not exist on ResNet-20-R. This motivates analysts to inspect those two groups of models from ResNet-20-R \texttt{2023} and ResNet-20-NR \texttt{2023}, to explore and investigate their local structure.
They notice that ResNet-20-R has ``flat'' loss landscapes in comparison to ResNet-20-NR (Figure~\ref{fig:resnet_case_study}.2) (\textbf{T2.1}). This is confirmed in the persistence diagram view since ResNet-20-NR has larger fluctuation than ResNet-20-R (Figure~\ref{fig:resnet_case_study}.3), and the merge tree of ResNet-20-NR shows there are many components treated as local minimum merged, and the other merge tree of ResNet-20-NR shows there is only one minimum in the loss landscape (Figure~\ref{fig:resnet_case_study}.4). This means \emph{the ResNet-20 architecture can result in better loss landscape both globally and locally with the residual connection (\textbf{T2.2}).}

\subsection{Exploring PINN Failure Modes}
\label{sxn: pinn}
Recent research brought the so-called Physics-Informed Neural Network (PINN) to much attention in the scientific machine learning community. The idea of a PINN is to incorporate physical domain knowledge into the machine learning models to solve different physical problems. Krishnapriyan et al.~\cite{krishnapriyan2021characterizing} demonstrate that such a PINN model performs well on relatively trivial problems while failing drastically on more complex problems. They further indicate that all of these failure modes are related to the setup of the PINN, which makes the loss landscape difficult to optimize. Thus, the motivation of this case study is to diagnose PINN models when they fail through our framework.

In this case study, we consider a one-dimensional convection problem, a hyperbolic partial differential equation that is commonly used to model transport phenomena~\cite{krishnapriyan2021characterizing}:

\begin{align}\label{eq:PINN_1}
    \frac{\partial u}{\partial t} + \beta \frac{\partial u}{\partial x} = 0, x \in \Omega, t \in [0, T],
\end{align}
 \vspace{-4mm}
\begin{align}\label{eq:PINN_2}
    u(x, 0) = h(x), x \in \Omega.
    \vspace{-4mm}
\end{align}
where $\beta$ is the convection coefficient and $h(x)$ is the initial condition. The general loss function for this problem is
 \vspace{-1mm}
\begin{align}\label{eq:PINN_3}
    L(\theta) = \frac{1}{N_{u}}\sum_{i=1}^{N_{u}} (\hat{u}-u_0^i)^2 + \frac{1}{N_{f}}\sum_{i=1}^{N_{f}} \lambda_i (\frac{\partial \hat{u}}{\partial t} + \beta \frac{\partial \hat{u}}{\partial x})^2 + L_B
    \vspace{-4mm}
\end{align}
where $\hat{u} = N N (\theta, x, t)$ is the output of the NN, and $L_B$ is the boundary loss. Here, our analysts apply the PINN’s soft regularization to this problem and optimize the loss function to explore how the error and loss change for different values of $\beta$. We train two separate models ($\beta=1$, $\beta=50$) using 100 different random seeds (from \texttt{1} to \texttt{100}), yielding $200$ models in total. As shown by Krishnapriyan et al.~\cite{krishnapriyan2021characterizing}, increasing $\beta$ results in a harder convection problem to solve, and this difficulty can be linked to changes in the corresponding loss landscape, which becomes increasingly complicated, such that optimizing the model becomes more complex.

\vspace{0.5mm} \noindent \textbf{Global Analysis}.
At first glance, the analysts notice that the models trained with $\beta=1$ are closer to each other than those trained with $\beta=50$ (Figure~\ref{fig:pinn_case_study}.1). 
After filtering out high mode connectivity edges, the analysts still see many curved lines connecting the models trained with $\beta=50$ (Figure~\ref{fig:pinn_case_study}.2), indicating that \emph{the higher value of $\beta$ results in more poorly connected minima (\textbf{T1.1})}.
Looking at model performance, the error rate is also higher for the higher value of $\beta$ (Figure~\ref{fig:pinn_case_study}.3, \textbf{T1.2}).
The analysts hypothesize that increasing $\beta$ leads to the model having a globally poorly connected and locally sharp loss landscape.
To verify this hypothesis, the analysts select two models with different values of $\beta$ for further local structure inspection. After filtering the mode connective value using the slider in the global view setting panel, the analysts select models \texttt{041} and \texttt{090}, since the former are well connected to other models, and the latter are poorly connected to other models.
Upon hovering over a specific node, all connected paths (within the specified range of mode connectivity values) are highlighted.

\vspace{0.5mm} \noindent \textbf{Local Analysis}.
Looking at their local structures, it turns out that the loss landscapes of the two models have many differences. The one with lower $\beta$ has a ``flat'' loss landscape and the other one ($\beta=50$)'s minimum is not very clear (Figure~\ref{fig:pinn_case_study}.4) (\textbf{T2.1}). However, the persistence diagram shows that the terrain of $\beta=50$ around the minimum is very rough and complex (Figure~\ref{fig:pinn_case_study}.5). It is also obvious from the merge tree that the number of branches of $\beta=50$ is much larger than those of $\beta=1$ (Figure~\ref{fig:pinn_case_study}.6) (\textbf{T2.2}), thus demonstrating that the increase of $\beta$ can lead to a poorly-connected, locally sharp loss landscape. This indicates that, \emph{when physical parameter $\beta$ increases, PINN model becomes harder to optimize as the learned models will rest separately on the global landscape, where each model is different from the other and all look sharp on the local landscape, which can correspond to the phase I in those five phases as described in Figure~\ref{fig:loss-landscape-types}.}

\section{Expert Interviews}

\tabulinesep=1.2mm
\begin{table}[tb]
  \caption{
    Questions designed for expert interview.
  }   
  \label{tab:expert_interview}
  \scriptsize
  \centering
\begin{tabu} {|X[1.23]| X[0.26] | X[0.16] |}
    \hline
    \textbf{Questions} & \textbf{Results} & \textbf{Aspect} \\
    \hline
    Rating: \textbf{\SYSNAME} helped me understand the model’s loss landscape. & $\mu = 4.3$, $\sigma = 0.24$ & \textbf{MD}  \\
    \hline
    Rating: \textbf{\SYSNAME} helped infer the model’s implicit properties. & $\mu = 3.4$, $\sigma = 0.37$   & \textbf{MD}\\
    \hline
    Rating: \textbf{\SYSNAME} provides a more efficient and effective workflow compared to traditional methods. & $\mu = 4.2$, $\sigma = 0.24$  & \textbf{MD}\\
    \hline
    Ranking: Please rank the views based on the attention you gave them during your exploration with \textbf{\SYSNAME}. & See main text   & \textbf{SUA}\\
    \hline
    Rating: Integrating my case studies into \textbf{\SYSNAME} was hassle-free.  & $\mu = 4.4$, $\sigma = 0.58$ & \textbf{SUA}
    \\
    \hline
    Free-form: What insights did you gain from using \textbf{LossLens}? Describe your experiences and specific instances where it provided valuable insights.& See main text & \textbf{PI}\\
    \hline
    Free-form: We have designed 3 visual representations in the Global Structure View. Which design is intuitive for representing the mode connectivity? & See main text & \textbf{VDI} \\
    \hline
    Rating: The visualization designs of \textbf{\SYSNAME} are easy to understand.  & $\mu = 3.1$, $\sigma = 0.19$  & \textbf{VDI}\\
    \hline
    Please provide any feedback or suggestions for improvements in \textbf{\SYSNAME}.  & See main text & \textbf{SI} \\
    \hline
\end{tabu}
\end{table}

All case studies and the system design were directly developed with our co-authors.
To further evaluate our framework, we organized a group interview involving five independent domain experts (referred to as E0 to E4). As our framework is tailored for researchers that focus on loss landscape research, we recruited E0 and E1 who actively focusing on loss landscape research, and E2, E3, and E4 who are researchers with backgrounds in deep learning, loss function research, and topological data analysis. The interview commenced by providing an overview of the framework's background and a comprehensive presentation of the system's functions. We demonstrated the analytical workflow using our two case studies, and the experts then explored the two case studies within our system. Finally, we let experts complete the evaluation questionnaire, including qualitative (free-form questions) and quantitative feedback (5-point scale rating questions). The whole process lasted 90 minutes.

\subsection{Questionnaire Design}
We designed our questionnaire based on five aspects:

\begin{itemize}
    \item \textbf{Model Diagnosis (MD):} Assessing how effectively \textbf{LossLens} helps in inferring the loss landscape and the model's implicit properties.
    \item \textbf{System Usage Analysis (SUA):} Evaluating \textbf{LossLens}'s practical aspects, including ease of use and backend integration.
    \item  \textbf{Patterns and Insights (PI): } Gathering your insights and the knowledge you derived from using \textbf{LossLens}.
    \item  \textbf{Visual Designs and Interactions (VDI):} Gauging your experience with the tool's visual and interaction design.
    \item \textbf{Suggestions and Improvement (SI):} An open-ended section for you to provide any additional feedback or suggestions for the tool.
\end{itemize}

Table~\ref{tab:expert_interview} provides the survey questions and summary statistics from our participants.

\subsection{Evaluation Results}
Overall, the framework design received positive feedback, with an overall mean rating of 3.88 (SD=0.63). Top-ranked visualizations were: global structure view (mean rank 1.75, SD=0.96), merge tree (mean rank 2.50, SD=1.29), and loss contour (mean rank 3.00, SD=0.81).
\begin{figure}[t!]
\centering
\includegraphics[width=0.80\columnwidth]{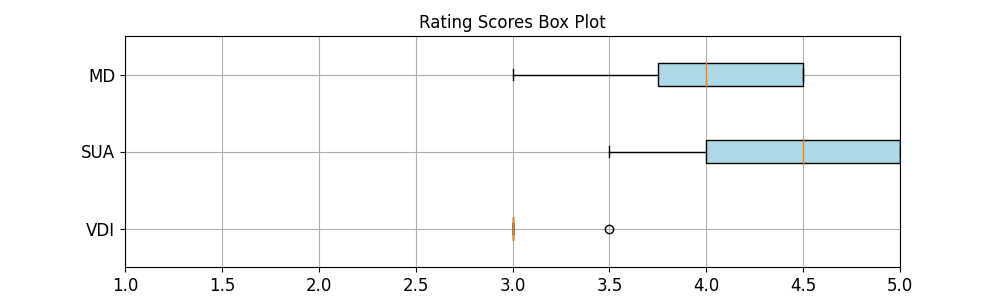}
      \caption{Rating distribution grouped based on aspects.}
    \label{fig:result}
\end{figure}
Here we provide the statistical insights from each aspect we designed based on the questionnaire results. According to Figure~\ref{fig:result}, the feedback from the users on the system reveals that Model Diagnosis (MD) was consistently well-received, with a tight clustering of high scores indicating general satisfaction. In contrast, System Usage Analysis (SUA) presented a broader range of opinions, although still leaning positive, but included a notable outlier suggesting a unique user dissatisfaction that might need attention. This is due to some engineering issues in the backend, and we plan to make the system more compatible with different models. Visualization Design and Interaction (VDI) received a singular critical score, highlighting it as a potential area for improvement. We discovered that even though we tried to make the concept of \textbf{\SYSNAME} to be well-explained, there are still learning curves for new learners. As such, we have developed more detailed documentation for using the system.
\tabulinesep=1.2mm
\begin{table}[tb]
  \caption{
    Visualization Rankings
  }
  \label{tab:ranking}
  \scriptsize
  \centering
\begin{tabu} {|X[0.5]| X[0.3] | X[0.3] |}
    \hline
    \textbf{Visualization} & \textbf{$\mu_{rank}$} & \textbf{$\sigma_{rank}$} \\
    \hline
    Global Structure View & 1.75 & 0.957427 \\
    \hline
    Local Contour & 3.00   & 0.816497 \\
    \hline
    Merge Tree & 2.50  & 1.290994 \\
    \hline
    Persistence Diagram & 3.75 & 1.258306 \\
    \hline
\end{tabu}
\end{table}
As shown in Table.~\ref{tab:ranking}, we received high rankings on our designed global structure view and local structure view. This indicates the users are more engaged in such design which helps them discover more information easily.

Comments include: 
E1, \emph{``Here is an example that I imagined this tool can be of help. When I tried to debug the model’s failure, the common practice was to change one hyperparameter and observe the performance change. By using the \textbf{\SYSNAME}, we can observe the changes in more features. If the change is conducted by increasing the model size, and the performance improvement is mild, this could be explained by one of the loss landscape features, e.g. the connectivity is not improved as well, indicating that the model size is large enough. Then we can try another hyperparameter such as an optimizer. It is known that some loss landscape properties are more correlated with specific hyperparameters, so it is useful to aggregate the local and global structures in one software.
(PI)''};
E0, \emph{``It is good to see an integrated pipeline built up for loss landscape research and model diagnosis in a different way. Usually, when we are trying to visualize the loss landscape, we only use a Jupyter Notebook to plot the Hessian-based two-dimensional contour. With this tool, I think it could be much easier to investigate a complicated model by running this pipeline (SUA).''}
Suggestions for improvement include: E3, \emph{``I really see the value of the framework, but to make the system more extendable, we can keep working on the scalability of the pipeline so that those visualizations can be generated efficiently. It is more of an engineering problem we need to consider in the future. (SI)''}.
Full feedback from our interviews can be found in the supplemental material.

\section{Discussion and Conclusions}
\label{sec:discussion}

In this work, we proposed a visual analytics framework, \textbf{\SYSNAME}, to help machine learning practitioners explain, diagnose, and validate deep learning models through loss landscape analysis. 
We employ D3.js and React to build our user interfaces, and we use PyTorch, loss-landscapes\footnote{https://github.com/marcellodebernardi/loss-landscapes}, Topology ToolKit, Flask, and MongoDB to compute and store the data for visualization. 
The code is available on GitHub\footnote{https://github.com/Vis4SciML/loss-lens}. 
Overall feedback was positive; however, some limitations were identified.

\vspace{1mm}
\noindent \emph{Computational Scalability}. Computing all the necessary data to be visualized is time-consuming and depends on (1) the number of pre-trained models, (2) the number of parameters in the models (e.g., for metrics like CKA similarity), (3) the ``range'' of the loss landscape (i.e., how far in each direction the neural network is perturbed), and (4) the ``resolution'' of the loss landscape (i.e., the number of model evaluations). 
We trained models and computed all metrics offline, since they require both long compute times and large memory overheads.
For instance, generating all necessary data for local structures took approximately 4 hours (2 hours for 200 PINN models and 2 hours for 8 ResNet models) on an M1 MacBook Pro with a specific environment configuration. Generating the global structure data required more GPU-intensive resources, which took approximately 4 hours (2 hours for PINNs and 2 hours for ResNets) on a single node (1 AMD Milan processor and 4 NVIDIA A100 GPUs) of the Perlmutter supercomputer at the National Energy Research Scientific Computing Center (NERSC). We believe that optimizing these algorithms, for example through parallelization, could significantly improve computational performance and reduce processing times.
While LossLens can only provide an overview of a finite number of trained models, we discussed the design requirements with our domain experts, and they confirmed that 3-10 models are usually sufficient for global structure analysis, especially since the metrics are measured pairwise.
To verify this, we trained 100 models in our second case study. Importantly, we were able to draw similar conclusions compared to cases with only 4 models. While interesting, we reiterate that since the global metrics are measured pairwise, similar conclusions can be drawn from only a handful of models. Nevertheless, we leave this decision to the user by allowing them to upload any number of models.

\noindent \emph{Visualization Scalability}.
Regarding visualization scalability, we acknowledge that the current encoding method may lead to overlap issues as the number of trained models increases. However, our domain experts prefer the current encoding format primarily because it effectively represents the concept of local and global information. For instance, the local smoothness of the loss landscape is derived from individual trained models, so it makes sense to encode this information locally (i.e., surrounding each node). Similarly, the representations of mode connectivity (depicted as curves) and CKA similarity (encoded in the spatial layout) are well-suited for conceptualizing the pairwise relationships that underlie the global structure.
In terms of sampling density, our domain experts assert that a finite number of points is sufficient, and indeed, insights into model properties can be obtained from just a few strategically trained models. However, to ensure that our system can be scaled up to any number of models, we also take steps to ensure that our visualizations make sense as the number of models grow. For example, to mitigate potential issues like visual overlap, we implement additional interactions including zooming and panning, toggling of specific components, label visibility controls, etc.
These features allow users to dynamically adjust the visualization to focus on areas of interest and reduce visual clutter when necessary. 

\vspace{1mm}
\noindent \textbf{Future Work.}
In the future, we hope to explore how we can visualize the loss landscape using higher-dimensional projections of the loss function. Such a method might be useful to reveal more insights and contribute to both visualization and machine learning communities. We also hope to extend our framework to support more state-of-the-art deep learning models.

\section{ACKNOWLEDGMENTS}
This work was supported by the U.S. Department of Energy, Office of Science, Advanced Scientific Computing Research (ASCR) program under Contract No. DE-AC02-05CH11231 to the Lawrence Berkeley National Laboratory and Award Number DE-SC0023328 to Arizona State University (“Visualizing High-dimensional Functions in Scientific Machine Learning,” program manager Margaret Lentz).
This research used resources of the National Energy Research Scientific Computing Center (NERSC), a Department of Energy Office of Science User Facility using NERSC award ASCR-ERCAP0026937.

\def\refname{REFERENCES}

\appendix 
\newpage
\begin{figure*}
    \centering
    \includegraphics[width=1\columnwidth]{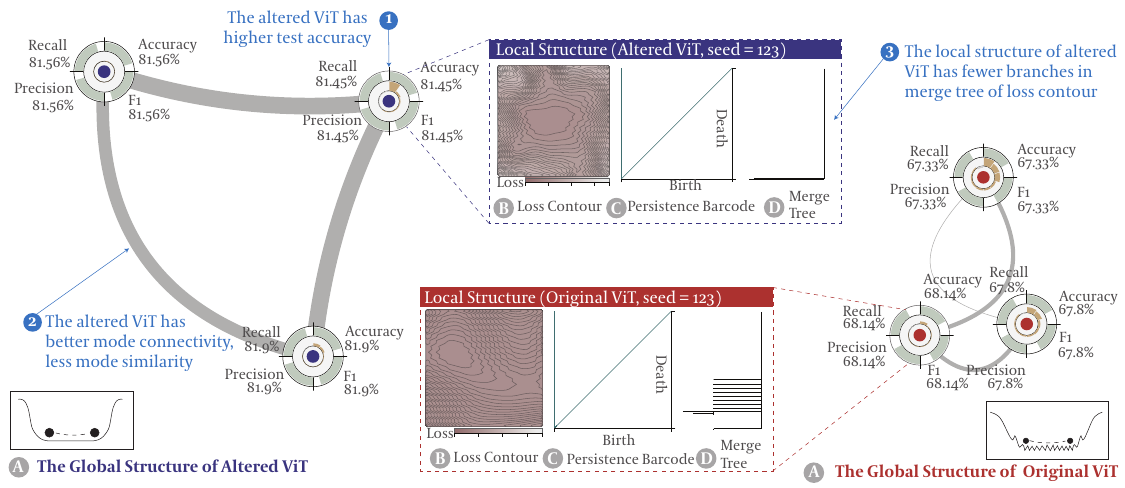}
    \vspace{-2mm}
    \caption{
    Demonstration of our \textbf{\SYSNAME} visual analytics system to diagnose a Vision Transformer (ViT) and the effect of data alteration.
    (1) The models (multiple variants of the same model due to the stochasticity) of an altered ViT (trained with CIFAR10-C, including common image corruptions and perturbations)
    exhibit higher test accuracy than those of the original ViT (trained with CIFAR10).
    The models are obtained using the same model architecture but different weight initializations and training stochasticity.
    (2) The altered ViT has better mode connectivity between models than the original ViT.
    This suggests that the loss landscape of the altered ViT has a more globally connected structure than the original ViT.
    However, the similarity between models is lower than those of the original ViT, suggesting that the trained altered ViT models exhibit a higher degree of variance than their original ViT counterparts.
    (3) The merge tree of model \texttt{123} of the altered ViT has fewer branches than the original ViT, confirming that the altered ViT has a flatter local minimum and, consequently, performs better.
    In contrast, the original ViT performs relatively poorly due to the limited size of the dataset.
    }
    \label{fig:teaser}
\end{figure*}

\section{Data Alteration of Vision Transformer}
\label{sxn:vision-transformer}

In our first case study, we explore how data alteration can influence the loss landscape. As the size of machine learning models continues to grow, with some models now encompassing billions of parameters, the challenge of insufficient training data becomes increasingly apparent. To address this issue and enhance the models' generalizability, methods such as data augmentation and adversarial training have been introduced. Those techniques help create a more varied and representative dataset, which in turn allows the model to learn more robust and generalized features, ultimately improving its performance on unseen data. Thus, in this case study, we verify that our framework is able to diagnose the data diversity that improves the model, which can be captured by our designed visualizations. To mimic improving the model through data, we demonstrate how our framework supports diagnosing the influence of using the same amount of CIFAR10 and CIFAR10-C images separately to train the Vision Transformer (ViT). 
We employ the standard architecture of the ViT model, which is configured using a patch size of four, hidden embedding dimension equal to 1024, MLP dimension of 2048, 16 heads, and a depth of six blocks. 
We use standard CIFAR10 and CIFAR10-C to train two models respectively.
To make the comparison consistent, we randomly sampled the CIFAR10-C dataset so that the sampled training dataset has the same size as the CIFAR10 dataset.
To help distinguish them,  we name the former as ``original ViT'' and the latter as ``altered ViT''. In this case study, we select three arbitrary random seeds (\texttt{123}, \texttt{123456}, and \texttt{2023}) to train the models\footnote{We pick arbitrary numbers to avoid random seed optimization.}, such that each model has three models based on those random seeds.

\vspace{2mm} \noindent \textbf{Global Analysis}.
To begin with, the analysts notice that the two models have different structures from the global structure view. By looking at the outer rings of the circle (Figure~\ref{fig:teaser}.1), the models of the altered ViT have higher test accuracy than those of the original ViT, and the altered ViT has better mode connectivity between models than the original ViT (Figure~\ref{fig:teaser}.2). This indicates that \emph{the loss landscape of the altered ViT seems to have a globally better-connected loss landscape than the original ViT (\textbf{T1.1})}. However, the similarity between models is larger than those of the original ViT (Figure~\ref{fig:teaser}.2), indicating that \emph{the trained altered ViT models tend to have a larger variance than the trained original ViT models (\textbf{T1.2}) in its prediction}.
Indeed, CIFAR10-C has different corruption data based on CIFAR-10, which could lead to the fact that the trained models perform differently.
Relatively, the global analysis indicates that the altered ViT has a \emph{locally flat, globally well-connected landscape, where models are less similar (Figure~\ref{fig:loss-landscape-types}. IV-A)}, and the original ViT has a \emph{locally sharp, globally poorly-connected landscape (Figure~\ref{fig:loss-landscape-types}. I)}. According to our co-authors from the machine learning community, the interpretation of this can be the following: \emph{Training ViT on a small dataset is essentially overfitting, where all trained models are similar to each other and the connectivity is poor between them. Thus, increasing the data diversity only can mitigate that as shown in altered ViT, where models are getting more stable in terms of the Hessian spectrum as well as better mode connectivity between models, thus improving the model from phase I to phase IV-A. }

\vspace{2mm} \noindent \textbf{Local Analysis}.
The analysts are interested in what the local structure of these two models looks like. 
According to the radial bar charts of the inner ring of the circles, they notice that the model original ViT \texttt{123} has the flattest local area (small Hessian eigenvalues), while the altered ViT \texttt{123} has the sharpest local area (one big Hessian eigenvalue) among their group of models(Figure~\ref{fig:teaser}. A, \textbf{T2.1}). The analysts select them to compare their local structures to see how they differ from each other on the local scale.
By looking at their loss contour (Figure~\ref{fig:teaser}. B), they observed one local minimum for each model (\textbf{T2.2}).
The merge tree of model \texttt{123} of the original ViT has more branches than the altered ViT (Figure~\ref{fig:teaser}.3), which verified that the latter has a flatter local minimum and thus performs better (\textbf{T2.2}).

\end{document}